\documentclass[10pt,twocolumn,letterpaper]{article}

\usepackage{wacv}              

\usepackage{graphicx}
\usepackage{amsmath}
\usepackage{amssymb}
\usepackage{booktabs}
\usepackage{balance}

\usepackage{multirow}
\usepackage{enumitem}

\usepackage[pagebackref,breaklinks,colorlinks,bookmarks=false ]{hyperref}
\usepackage{xcolor}
\usepackage{colortbl}  

\usepackage[capitalize]{cleveref}
\crefname{section}{Sec.}{Secs.}
\Crefname{section}{Section}{Sections}
\Crefname{table}{Table}{Tables}
\crefname{table}{Tab.}{Tabs.}

\def\wacvPaperID{386} 
\def\confName{WACV}
\def\confYear{2024}

\begin{document}

\title{Few-shot Shape Recognition by Learning Deep Shape-aware Features}

\author{Wenlong Shi$^{\star,\triangledown}$, Changsheng Lu$^{\star,\Diamond}$, Ming Shao$^{\dagger,\S}$, Yinjie Zhang$^{\triangledown}$, Siyu Xia$^{\dagger,\triangledown}$, Piotr Koniusz$^{\dagger,\Diamond,\clubsuit}$ \\
$^{\triangledown}$School of Automation, Southeast University \quad 
$^{\Diamond}$The Australian National University \\
$^{\S}$University of Massachusetts, Dartmouth \quad
$^\clubsuit$Data61/CSIRO\\
{\tt\small \{wenlong.shi, xsy\}@seu.edu.cn, \{changsheng.lu, piotr.koniusz\}@anu.edu.au, mshao@umassd.edu}
}
\maketitle

\def\thefootnote{$\star$}\footnotetext{Co-first author, $^\dagger$co-correspondence. $^{\S}$This material is based upon work supported by the National Science Foundation under Grant No. 2144772.}

\begin{abstract}
  Traditional shape descriptors have been gradually replaced by convolutional neural networks due to their superior performance in feature extraction and classification. The state-of-the-art methods recognize object shapes via image reconstruction or pixel classification. However, these methods are biased toward texture information and overlook the essential shape descriptions, thus, they fail to generalize to unseen shapes. We are the first to propose a few-shot shape descriptor (FSSD) to recognize object shapes given only one or a few samples. We employ an embedding module for FSSD to extract transformation-invariant shape features. Secondly, we develop a dual attention mechanism to decompose and reconstruct the shape features via learnable shape primitives. In this way, any shape can be formed through a finite set basis, and the learned representation model is highly interpretable and extendable to unseen shapes. Thirdly, we propose a decoding module to include the supervision of shape masks and edges and align the original and reconstructed shape features, enforcing the learned features to be more shape-aware. Lastly, all the proposed modules are assembled into a few-shot shape recognition scheme. Experiments on five datasets show that our FSSD significantly improves the shape classification compared to the state-of-the-art under the few-shot setting. 
\end{abstract}

\section{Introduction}
%
Shape recognition has been a critical area of research in computer vision, with applications in numerous fields, such as industrial automation \cite{yu2022capri,lu2018viewpoint}, botanics classification \cite{swedishleaf}, and fine-grained shape recognition for medical organs \cite{yang2020circlenet}. In industrial automation, it is usually required to identify the textureless components that possess a unique shape, \eg the workpiece classification. In botany science, diverse herbs collected in the wild need to be classified and made taxonomy. Moreover, the shape recognition is useful to identify lesions or pathological changes in medical diagnosis \cite{yang2020circlenet}.


As demonstrated in the variety of recognition tasks,  convolutional neural networks (CNN) allow high-dimensional image data to be presented as semantic features. Such representations capture both shape and texture information. While existing studies argue that shape information plays dominant roles in general recognition tasks\cite{ritter2017cognitive,lecun2015deep,kubilius2016deep,zeiler2014visualizing}, many studies highlight that local textures provide adequate information for various vision tasks\cite{brendel2019approximating,geirhos2018imagenet,ballester2016performance,funke2017synthesising}. There are ongoing efforts to combine texture and shape information \cite{kong2021competition,hwang2021stfnet}. Nonetheless, the majority of works focus on utilizing \textit{texture information} to capture object shapes in object segmentation \cite{long2015fully,ronneberger2015u}, arbitrary-shaped text detection \cite{cai2022arbitrarily,zhu2021fourier,qi2022ktext,wang2021pan++}, salient object detection \cite{zhao2019egnet,wei2020label,qiao2021salient,mukherjee2017salprop,wu2019stacked}, circle/ellipse detection \cite{wang2022eldet,lu2019arc,lu2017circle}, and shape classification \cite{zhang2021scn,atabay2016convolutional,atabay2016binary,kurnianggoro2017shape}. However, these texture-based methods do not target extracting generic shape information suitable for shape recognition of common textureless objects or novel textureless objects.

Finding discriminative representations for object shapes is non-trivial, and no ultimate mathematical definition has ever been established so far. Traditional shape descriptors mathematically approximate the geometric information of shapes and demonstrate their efficacy empirically\cite{sonka2014image}. These methods enable fast computation while maintaining high accuracy. Moreover, explicit mathematical definitions produce interpretable descriptors, and therefore, are still widely used today.

Zeiler \etal \cite{zeiler2014visualizing} argue that CNN features carry limited shape information. To address this issue, the approaches proposed by \cite{kervadec2021beyond,ren2021ship,li2020ggm,hao2018ami,singh2022learning} attempt to combine dedicated shape descriptors with well-established CNN features so that the learned representation can carry more shape information. These methods provide some degree of interpretable modeling and representations. 
Nonetheless, these methods model semantic and shape representations separately, \ie, the feature integration is applied right before output layers without further coupling in earlier layers.

In this paper, we formulate a generic shape representation for few-shot shape recognition. In contrast to existing representations, our model couples the shape descriptions with  deep learning for few-shot learning. Our FSSD model is based on a \textit{matching network} for few-shot learning, but with three unique contributions. Firstly, we propose to use group equivariant convolutional neural networks (G-CNN) instead of regular convolutional networks to help the embedding module handle rotations of input. Secondly, to encourage networks to \textit{learn} interpretable shape features comparable to conventional \textit{hand-crafted} shape descriptions, we propose novel shape primitives learned through dual attention architecture. The shape primitives serve as the basis to describe various shapes and help ``explain out'' the formulation of each input shape through reconstruction and visualization. Thirdly, we incorporate supervision through pairwise decoders to align the original and reconstructed shape features. This ensures the fidelity of shape primitives and the efficacy of the learned shape representation.




\vspace{0.2cm}
The main contributions of our work are as follows:\vspace{-0.2cm}
\renewcommand{\labelenumi}{\roman{enumi}.}
\begin{enumerate}[leftmargin=0.6cm]
    \item To our best knowledge, we propose the first few-shot shape recognition model capable of recognizing seen and unseen geometric shapes.\vspace{-0.2cm}
    \item We propose a dual attention architecture to learn shape primitives to improve shape representation learning. We add supervision through shape masks and the edge of objects to guide shape-aware feature learning. \vspace{-0.2cm}
    \item We collect a dedicated shape recognition dataset and transform an existing image dataset into shapes. We demonstrate that the proposed approach outperforms the state-of-the-art on five datasets.
\end{enumerate}

\section{Related works}
\subsection{Shape Descriptors} 
Defining the shape of an object is a challenging task, and current methods for shape descriptions generally focus on the following factors \cite{sonka2014image}: i) input representation (\ie, the description of an object can be based on its boundary or on its whole region); ii) ability to reconstruct the object; iii) ability to recognize incomplete shapes, 
and iv) robustness to various transformations such as translation, rotation, and scale. Therefore, most shape description methods can be classified into two categories: i) boundary-based methods and ii) region-based methods. Boundary-based methods include Hough transform \cite{hough1959machine,duda1972use,atherton1993using,murillo2012implementation,le2010real,oh2015extending,kimura2002extension,rababaah2020angle}, Fourier descriptors \cite{zahn1972fourier,gode2016object}, curvature scale-space descriptors \cite{mokhtarian1996robust,jalba2006shape}, shape context \cite{belongie2002shape}, segment boundary descriptors \cite{KoniuszM10}, wavelet transform descriptors \cite{chuang1996wavelet} and more  \cite{yang2020learning,landesa2010fast,moon2002optimal,wang2001detection,geiger2018quantum,kanatani2004automatic,wu2007shape}. Region-based methods include angular radial transformation \cite{bober2001mpeg,xu2019smart,jun2015led}, image moment \cite{flusser1993pattern}, and general Fourier shape descriptors \cite{zhang2002shape}.


Unlike existing works, this paper presents a deep learning pipeline formed from the perspective of shape description. Such a network can learn boundary-based generic shape features that are robust to translations and rotations, and can reconstruct object shapes from the basis set. 

\begin{figure*}[ht]
  \centering
  \includegraphics[width=\linewidth]{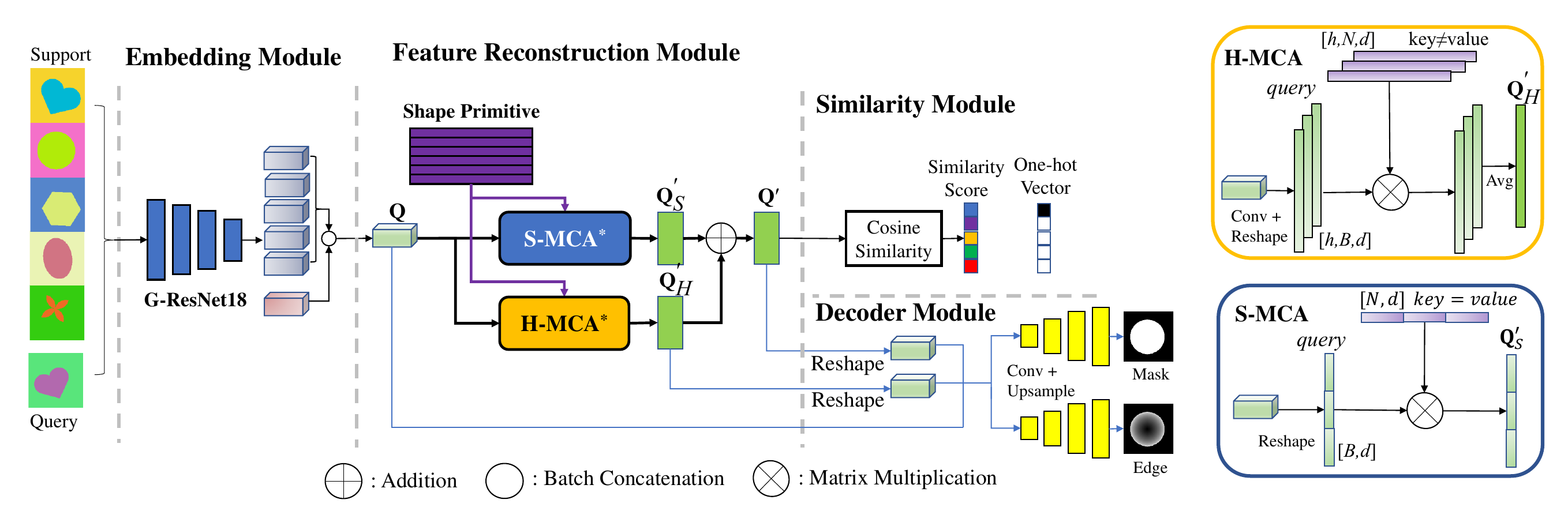} 
  \caption{
  The overall architecture of the proposed FSSD consists of four modules (from left to right): embedding module, feature reconstruction module, similarity module, and decoder module. FSSD follows a few-shot learning pipeline, where feature reconstruction module learns a set of primitives for shapes and reconstructs the output features of embedding module using the input support set and query set. Feature reconstruction module consists of Holistic Multi-head Cross-Attention (H-MCA) and Standard Multi-head Cross-Attention (S-MCA). The pairwise decoders enables supervision through ground truth shape masks and edges to align original and reconstructed shape features. The similarity is calculated using the reconstructed support set features and query set features, enabling shape classification.}
\label{fig:pic1}
\end{figure*}

\subsection{Few-shot Learning (FSL)} 
FSL can perform image classification, object detection, or segmentation given limited training data. Existing FSL works can be divided into two categories: i) metric-based methods (\eg, Prototypical Networks, Relation Networks); and ii) optimization-based methods such as MAML. Relation Networks \cite{sung2018learning} introduce a relation module to learn the similarity between the features of two input samples. Prototypical Networks \cite{snell2017prototypical} map a set of samples per class into a prototype vector. Then, classification is performed by measuring the cosine similarity between query samples and prototypes. MAML \cite{finn2017model} trains models iteratively on multiple tasks. 

While existing FSL pipelines target various vision tasks, \eg, few-shot object detection \cite{fan2020few,kang2019few}, few-shot segmentation \cite{li2020fss,li2021adaptive}, and few-shot keypoint detection \cite{Lu_2022_CVPR,SalViT}, few-shot shape recognition and deep shape-aware features are still under-explored. We argue and show that good shape descriptors are essential for good performance compared to other factors, \eg, distance metric in similarity measurement. 

\subsection{Shape Recognition} 
Most shape recognition methods utilize simple CNNs to classify binary shape images \cite{zhang2021scn,atabay2016convolutional,atabay2016binary,kurnianggoro2017shape}. Some other works encode and map shapes into high-dimensional spaces for classification \cite{jetley2017straight,miksys2019straight}. In shape detection, special detection frameworks are used instead of bounding boxes. For example, Kang \etal \cite{kang2020bshapenet} proposed the use of bounding masks to regress the object edges, while ellipse detection \cite{wang2022eldet,dong2021ellipse} utilizes elliptical bounding boxes to detect the most common ellipse shapes in the natural world. Tasks such as arbitrary-shaped text detection and salient object detection are closely related to shape reconstruction. In arbitrary-shaped text detection \cite{cai2022arbitrarily,zhu2021fourier,qi2022ktext,wang2021pan++}, the aim is to capture regions around various text shapes rather than using bounding boxes. Salient object detection \cite{zhao2019egnet,wei2020label,qiao2021salient,mukherjee2017salprop,wu2019stacked} focuses on reconstructing more accurate object boundaries. 


This paper proposes a novel dual attention mechanism to learn a finite set of universal shape primitives. The learned set of primitives through known shapes can be extended to compose, represent, and interpret unseen shapes, enabling robust yet discriminant shape features for the classification of new shapes.

\section{Proposed method}
We develop a few-shot model which can focus  on shape characteristics and maintain a high accuracy of shape classification. The entire model consists of four parts: (i) embedding module, (ii) feature reconstruction module, (iii) similarity module, and (iv) decoding module. 

\subsection{Problem Statement}


This paper uses episode-based training \cite{vinyals2016matching} often used in few-shot learning. In each episode, the so-called $c$-way $m_1$-shot learning is applied. Specifically, $c$ classes are randomly selected from the training set, with $m_1$ samples randomly selected from each class to form the support set $\mathcal{S} = {(\mathbf{x}_i,y_i)}^{m_1\times c}_{i=1}$, and $m_2$ samples randomly selected from each class to form the query set $\mathcal{Q} = {(\mathbf{x}_j,y_j)}^{m_2 \times c}_{j=1}$.

A typical few-shot learning network consists of two parts: an embedding module and a similarity module. To accommodate the shape primitives and supervision, we introduce a novel feature reconstruction module and a decoding module. Firstly, the embedding module employs a group-equivariant convolutional network (G-CNN) that incorporates group transformations into the convolutional operations by rotation-equivariant operations. Secondly, a novel dual attention architecture is implemented to learn shape primitives for feature reconstruction. Thirdly, the similarity module uses the cosine similarity to calculate the similarity between the support set features and the query set features. Finally, the original features and the support/query reconstructed features are fed into the decoding module to recover the shape mask and edges. 
The overall architecture of our method is shown in Figure ~\ref{fig:pic1}. 


\subsection{Embedding Module} 
\label{sec:G-CNN}


Shape descriptors should be robust to geometric transformations such as translation, rotation, and scale changes.  
Recent works have shown that group-equivariant networks can capture objects better as they do not require learning separately each object pose. 
Given the transformation group $g$, group-equivariant networks \cite{cohen2016group} 
use the following operator:
\begin{equation}
  \left[T_g \mathbf{f} \right] \otimes \Psi=T_g[\mathbf{f} \otimes \mathbf{\Psi}],
  \label{equ:4}
\end{equation} where $\mathbf{f}$ is the feature map or image, $\mathbf{\Psi}$ is the convolution filter, $\otimes$ is the convolution, and $T_g$ indicates the proposed transformation groups in \cite{cohen2016group}, $p_4 \!=\!\{r,r^2,r^3,e\}$ and $p_4^m \!=\! \{e,r,r^2,r^3,mr,r^3m,rm, mr^3,r^2m,mr^2,rmr,m\}$, where $e$ represents an equivalent transformation, $r$ represents a counterclockwise rotation of 90$^{\circ}$ and $m$ represents a mirror transformation. Next, take the $p_4$ group as an example. Let $T_g \in p_{4}$, $\mathbf{f}$ be the image, and $\mathbf{\Psi}$ be a CNN filter. Then, the equivariance property means that applying $T_g$ to $\mathbf{f}$ followed by convolution filter $\mathbf{\Psi}$ is identical to applying filter $\mathbf{\Psi}$ first to $\mathbf{f}$ followed by $T_g$ in the feature domain. By using the equivariance property, backbone is ``invariant'' to transformations so it does not have to learn each object pose separately. By working in the equivariant feature domain rather than on the raw images, our similarity module then learns invariance on equivariant features by learning to match shapes in different poses. This process has been shown in 
Figure \ref{fig:pic1}.

\subsection{Feature Reconstruction Module}


Discovering the basis or principal components has been popular practice in signal processing and representation learning, \eg, Fourier transform, principal component analysis (eigenfaces), low-rank matrix analysis, sparse coding. Inspired by such works, we let shape descriptors be approximated by a finite set of shape primitives. We propose to learn a finite set of shape primitives $\mathbf{\Phi}$ to ground the shape features of the images for both training and unseen data. 

\vspace{0.1cm}
\noindent
\textbf{Attention-based Primitives.} The support \& query features in an episode are combined into a matrix $\mathbf{Q} \in \mathbb R^{B \times d}$, where $B$ is the batch size (\ie, the total number of support \& query images) and $d$ is the channel dimension. 

Firstly, the feature vector per image, extracted from G-CNN, is regarded as a single query. Each shape primitive is corresponding to one key-value pair. Once we obtain the query-key-value, the attention can be established. 
Similar to the conventional attention mechanism \cite{vaswani2017attention}, we have:
\begin{equation}
\begin{aligned}
\mathbf{W}&=\operatorname{softmax}\left(\frac{\mathbf{Q}\mathbf{\Phi}^T}{\sqrt{d_k}}\right), \\
  \mathbf{Q}'\!&=\mathbf{W}\mathbf{\Phi},
  \end{aligned}
  \label{equ:7}
\end{equation}
where $\mathbf{W}\in \mathbb{R}^{B\times N}$ contains attention scores, $d_k$ is a scaling factor, and $\mathbf{Q}' \in \mathbb{R}^{B \times d}$ contains reconstructed shape features. For any image with index $j$, the corresponding reconstructed shape feature vector $\mathbf{q}'_j$ can be written as:
\begin{equation}
  \mathbf{q}'_j = ~{\sum\limits_{i = 1}^{N}{w_{ji}\boldsymbol{\phi}_{i}}}.
  \label{equ:9}
\end{equation}
One can see that the reconstructed feature vector $\mathbf{q}'$ is obtained by a linear combination of shape primitives $\mathbf{\Phi}=[\boldsymbol{\phi}_1,\ldots,\boldsymbol{\phi}_N]$. This also demonstrates that using attention to model feature vectors with shape primitives is reasonable. In Section~\ref{sec:Visualization of shape primitives}, we empirically show that the linear addition of primitives results in continuously changing decoded images. 

\vspace{0.1cm}
\noindent
\textbf{Similarity Module.} The similarity score is computed for pairwise reconstructed support-query feature vectors. 
We use the cosine similarity as metric due to its simplicity. Given a pair of query and support samples, the similarity score $s$ is obtained by the so-called weighted sum-kernel:
\begin{equation}
  \begin{aligned}
    s &= \langle \mathbf{q}'_s,~\mathbf{q}'_q \rangle \\
    &= \bigg( {\sum\limits_{i = 1}^{N}w_{i}^{s}}\boldsymbol{\phi}_{i} \bigg)^\top\bigg( {\sum\limits_{j = 1}^{N}w_{j}^{q}}\boldsymbol{\phi}_{j} \bigg)\\
    &= \sum\limits_{i = 1}^{N}\sum\limits_{j = 1}^{N}w_{i}^{s}w_{j}^{q} \langle \boldsymbol{\phi}_{i},~\boldsymbol{\phi}_{j}\rangle.   \\
    \end{aligned}
  \label{equ:10}
\end{equation}
If $i=j$, $\langle \boldsymbol{\phi}_i, \boldsymbol{\phi}_j \rangle$ becomes larger than the case of $i\neq j$ (assuming the unit $\ell_2$ norm of vectors). Based on this fact and Eq.~\eqref{equ:10}, if two samples are indeed similar, the similarity function encourages larger weights on the same primitives from both samples.


\vspace{0.1cm}
\noindent
\textbf{Dual Attention Architecture.} In this section, we apply multi-head unit to extend single-head based shape primitives. Since we use 
%
$\mathbf{\Phi}$ 
as the ``basis'' to reconstruct inputs,  a common 
$\mathbf{\Phi}$ is learned and used. Standard Multi-head Cross-Attention (S-MCA) \cite{vaswani2017attention} will learn different sub-primitive matrices for each head and thus cannot achieve our aim. Thus, we propose a Holistic Multi-head Cross-Attention (H-MCA) that encourages a common 
$\mathbf{\Phi}$ across different heads. By ``holistic'' we mean maintaining the integrity of primitives. 



In the reminder of this paper, we use $\mathbf{Q}_H'$ to indicate the reconstruction via H-MCA. To that end, first, the dimensions of $\mathbf{Q}$ and $\mathbf{\Phi}$ (used as key) are mapped to $d' = d \times h$ dimension, while on the other hand $\mathbf{\Phi}$ (used as value) is duplicated $h$ times ($h$ is the number of heads). Thus, the output of each head is obtained from a linear combination of common primitives from set $\mathbf{\Phi}$, which preserves the structural integrity of the primitives, and helps   visualize primitives. However, H-MCA itself lacks flexibility in reconstructing the features $\mathbf{Q}$, sometimes leading to poor $\mathbf{Q}'$ due to inadequate shape characteristics. To accommodate and supplement H-MCA, S-MCA is added into H-MCA to achieve  enhanced reconstruction:
\begin{equation}
    \mathbf{Q}' = \text{H-MCA}(\mathbf{Q}) + \text{S-MCA}(\mathbf{Q}) = \mathbf{Q}'_H + \mathbf{Q}'_S,
    \label{eq:dualatt}
\end{equation}
where $\mathbf{Q}'_S$ is the reconstruction by S-MCA. We term Eq.~\eqref{eq:dualatt} as \textit{dual attention architecture}. One can easily see that the enhanced $\mathbf{Q}'$ satisfies Eq. \eqref{equ:9} while using supplementary term $\mathbf{Q}'_S$. This practice is also similar to residual structure in ResNet, where H-MCA basically reconstructs the identity feature $\mathbf{Q}'_H$, whereas S-MCA learns the subtle changes $\mathbf{Q}'_S$ to complete the reconstructed features $\mathbf{Q}'$.



\begin{figure}[t]
\centering
\includegraphics[width=.99\linewidth]{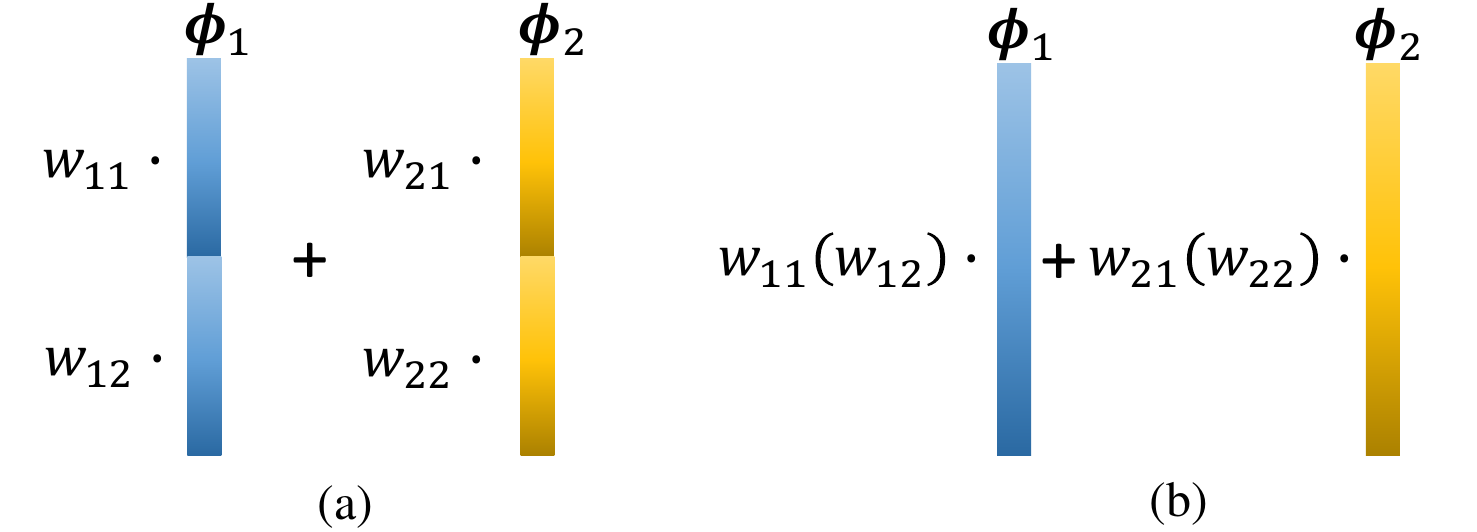}
\caption{
(a) Standard Multi-head Cross-Attention (S-MCA). Each primitive $\boldsymbol{\phi}_1, \boldsymbol{\phi}_2$ ($\mathbf{\Phi}=[\boldsymbol{\phi}_1, \boldsymbol{\phi}_2]$) is essentially divided into two pieces. (b) Holistic multi-head cross-attention (H-MCA). The primitives are complete regardless of the number of heads. In both cases, the number of primitives is 2, $h$=2, $B$=1, and attention tensor $\mathbf{W} \in \mathbb R^{2 \times 1 \times2}$.
}
\label{fig:S-MCAandH-MCA}
\end{figure}

Figure~\ref{fig:S-MCAandH-MCA}(a) shows an  example of how S-MCA works. The attention tensor $\mathbf{W} \in \mathbb{R}^{2\times 1 \times 2}$ assumes the number of primitives, heads, and batch size are 2, 2, and 1, respectively. Please note that S-MCA breaks the integrity of the primitives as the attention is performed on each head. Thus, if S-MCA is used alone, the reconstructed $\mathbf{Q}'$ is not obtained by the linear weighting of holistic primitives $\mathbf{\Phi}$, which contradicts Eq. \eqref{equ:9}.

\begin{figure}[t]
\centering
\includegraphics[width=1\linewidth]{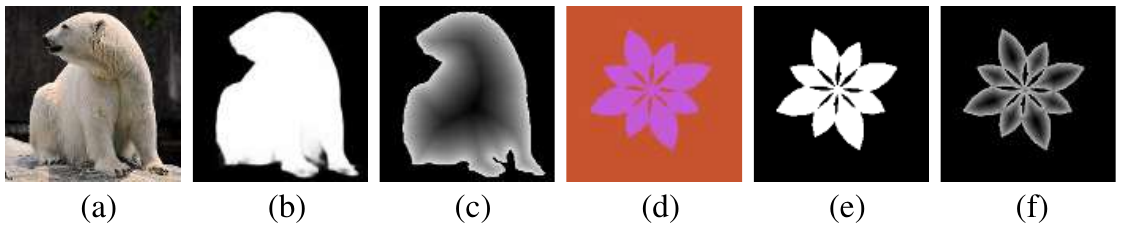}
\caption{
(a) shows a polar bear from the AwA2 dataset, and (d) shows an eight-petaled flower from the simple shape dataset. (a)(d) original image, (b)(e) mask image, and (c)(f) edge image.}
\label{fig:maskandedge}
\end{figure}

\subsection{Decoder Module}
\label{sec:Alignment of reconstructed features}
The reconstructed features $\mathbf{Q}'$ are used by the similarity module for  few-shot learning. $\mathbf{Q}'$ also requires additional constraints to achieve faithful reconstruction akin to  input features $\mathbf{Q}$. One might align $\mathbf{Q}'$ and $\mathbf{Q}$ via the Least Squares Error or KL-divergence. One might align both terms to enhance edges and the shape mask \cite{wei2020label}. In such a way, the shape supervision can be injected directly on top of the shape features $\mathbf{Q}$.

In our case, the alignment between $\mathbf{Q}$ and $\mathbf{Q}'$ is achieved through two decoders with common supervision: one decoder focuses on the edges, and the other focuses on the shape mask. Moreover, we encourage the similarity between $\mathbf{Q}'_H$ and $\mathbf{Q}$ as high as possible. 

By passing $\mathbf{Q},\mathbf{Q}'_H$ and $\mathbf{Q}'$ through two decoders, the decoders enforce their alignment as well as  learning of shape primitives. Both masks and edges (Figures~\ref{fig:maskandedge}(b)  and (c)) are used as ground truth, allowing the network to focus  on the shape information of  objects. Additionally, the decoder module also accomplishes the task of image reconstruction. 

\subsection{Loss Function}
Our loss function can be summarized as: 

\begin{equation}
    \mathcal{L} = ~\mathcal{L}_{cls} + ~\mathcal{L}_{res},
\label{equ:12}
\end{equation}
where $\mathcal{L}_{cls}$ and $\mathcal{L}_{res}$ are the loss functions for classification and image reconstruction, respectively. The loss function $\mathcal{L}_{res}$ includes six components:
\begin{equation}
    \mathcal{L}_{res} = \mathcal{L}_{mask}^{\mathbf{Q}} + \mathcal{L}_{edge}^{\mathbf{Q}} + \mathcal{L}_{mask}^{\mathbf{Q}'_H} + \mathcal{L}_{edge}^{\mathbf{Q}'_H} + ~\mathcal{L}_{mask}^{\mathbf{Q}'} + ~\mathcal{L}_{edge}^{\mathbf{Q}'},\nonumber
\label{equ:13}
\end{equation}
where $\mathcal{L}_{mask}$ and $\mathcal{L}_{edge}$ are the loss functions for mask and edge image reconstruction, respectively. $\mathcal{L}^Q$, $\mathcal{L}^{Q'_H}$ and $\mathcal{L}^{Q'}$ are the image reconstruction loss functions for the original features, holistically reconstructed features and dual reconstructed features, respectively. For the reconstruction loss function, MSE loss is used.

\begin{figure}[t]
\centering
\centering
\includegraphics[width=\linewidth]{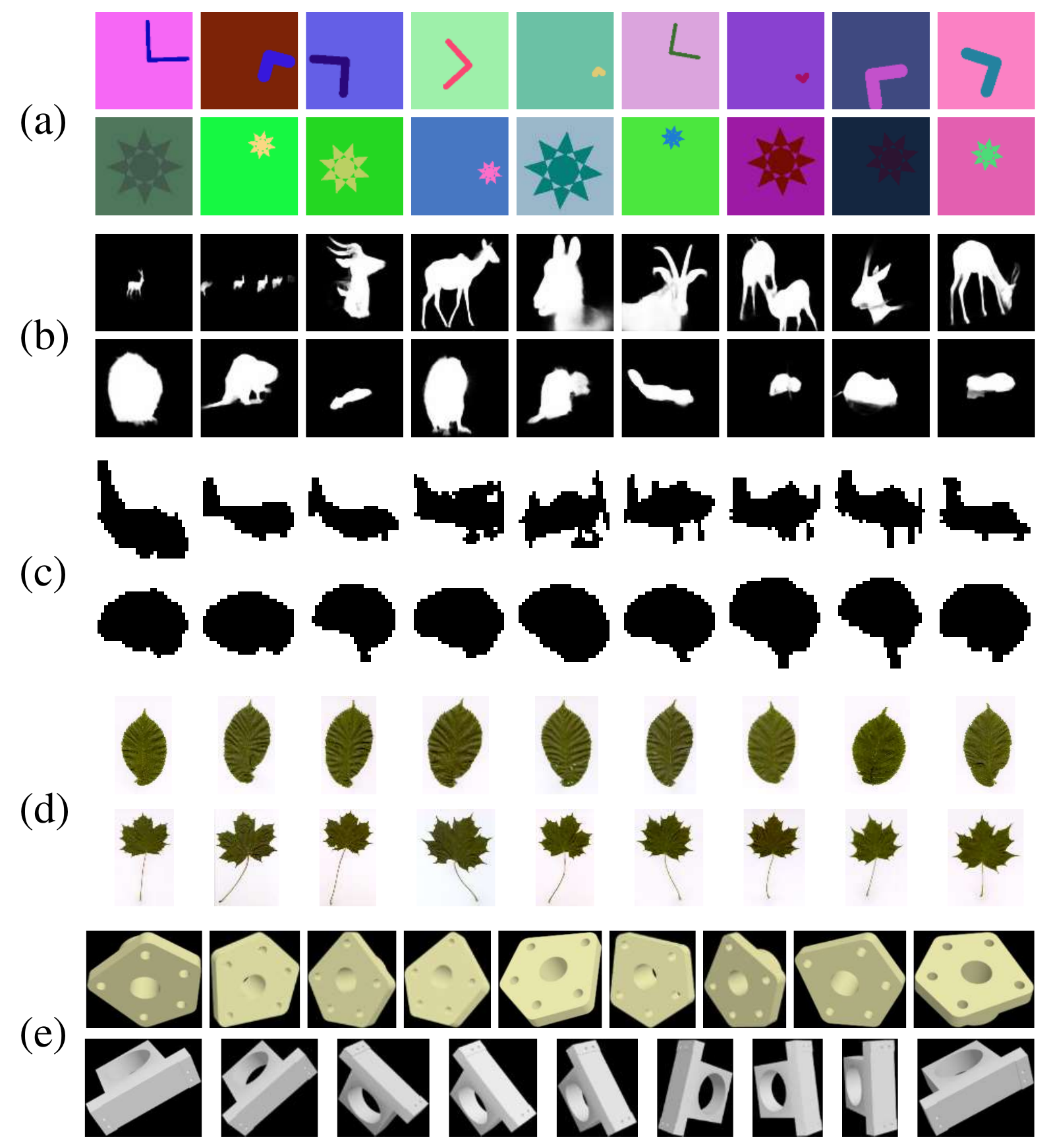}
\caption{
Example of shape images from \emph{five} datasets. Each row showcases the data from one class. (a) Simple shape dataset, \eg, \textit{folding line}, \textit{octagram}. (b) The shape-AwA2 dataset, including \textit{antelope}, \textit{beaver}. (c) Caltech 101 shapes with classes: \textit{airplanes side 2} and \textit{brain}. (d) Swedish leaf dataset. (e) Workpiece dataset.}
\label{fig:pic4}
\label{fig:datdaset}
\end{figure}
\begin{table*}[h]
\centering
\caption{
Comparison with traditional shape descriptors and classical few-shot learning methods across five datasets. The 5-way 1-shot and 5-way 5-shot results averaged over 5,000 test episodes are reported.}
\label{table:Algorithm comparison}
\resizebox{1\linewidth}{!}{
\begin{tabular}{lllcccccccccc}
\hline\noalign{\smallskip}
\multirow{2}*{Algorithm} & \multirow{2}*{Backbone} & \multirow{2}*{Params} & \multicolumn{2}{c}{Simple-Shape ($\uparrow$)} & \multicolumn{2}{c}{Shape-AwA2 ($\uparrow$)} & \multicolumn{2}{c}{Caltech 101 ($\uparrow$)} & \multicolumn{2}{c}{Workpiece shapes ($\uparrow$)} & \multicolumn{2}{c}{Swedish leaf ($\uparrow$)}\\
&&& 1-shot & 5-shot & 1-shot & 5-shot & 1-shot & 5-shot & 1-shot & 5-shot & 1-shot & 5-shot \\ \hline
Shape context\cite{belongie2002shape}  & - & 0M & 34.25$\%$ & 50.64$\%$ & 22.69$\%$ & 31.92$\%$ & 52.80$\%$ & 66.00$\%$ & 30.80$\%$ & 53.60$\%$ & 60.32$\%$ & 70.02$\%$\\
Fourier descriptors\cite{zahn1972fourier} & - & 0M & 55.76$\%$ & 65.60$\%$ & 18.49$\%$ & 23.77$\%$ & 38.17$\%$ & 48.91$\%$ & 46.57$\%$ & 62.53$\%$ & 38.09$\%$ & 54.96$\%$\\
Hu moment\cite{hu1962visual} & - & 0M & 80.05$\%$ & 87.49$\%$ & 21.85$\%$ & 22.72$\%$ & 47.74$\%$ & 50.63$\%$ & 47.32$\%$ & 50.16$\%$ & 47.73$\%$ & 64.44$\%$\\
ProtoNet\cite{snell2017prototypical} & ResNet18 & 43M & 83.90$\%$ & 80.79$\%$ & 23.66$\%$ & 34.07$\%$ & 66.28$\%$ & 85.41$\%$ & 67.84$\%$ & 86.55$\%$ & 47.56$\%$ & 91.01$\%$\\
RelationNet\cite{sung2018learning} & ResNet18 & 48M & 80.84$\%$ & 83.23$\%$ & 22.74$\%$ & 42.45$\%$ & 59.53$\%$ & 86.74$\%$ & 66.76$\%$ & 89.18$\%$ & 43.56$\%$ & 92.50$\%$\\
\rowcolor{green!20}
Ours & G-ResNet18 & 45M & \textbf{91.02}$\%$\ & \textbf{92.58}$\%$ & \textbf{43.35}$\%$ & \textbf{54.45}$\%$ & \textbf{80.19}$\%$ & \textbf{89.13}$\%$ & \textbf{97.29}$\%$ & \textbf{97.79}$\%$ & \textbf{88.58}$\%$ & \textbf{93.76}$\%$\\
\hline
\end{tabular}
}
\end{table*}

\begin{table}[t]
\caption{
Results for several backbones on the simple shape dataset.
}
\label{table:table1}
\begin{minipage}{0.48\hsize}
\resizebox{1.0\linewidth}{!}{
\begin{tabular}{llc}
Model  & Backbone & Acc ($\uparrow$)\\  \hline                
\multirow{4}{*}{Ours}  & 4-layer conv & 78.72$\%$ \\
& ResNet18 & 85.24$\%$\\
& ResNet50 & 85.83$\%$\\
& \cellcolor{green!20}G-ResNet18 & \cellcolor{green!20}\textbf{91.02}$\%$ \\ \hline
\end{tabular}
}
\end{minipage}
\begin{minipage}{0.48\hsize}
\resizebox{1.0\linewidth}{!}{
\renewcommand{\arraystretch}{1.3}
\fontsize{15}{15}\selectfont
\begin{tabular}{llc}
Model        & Backbone & Acc ($\uparrow$)   \\ \hline
\multirow{2}{*}{PrototNet}    & ResNet18   &   83.90$\%$   \\
                & G-ResNet18 &   86.59$\%$   \\ 
\multirow{2}{*}{RelationNet}    & ResNet18   &   80.84$\%$   \\
                & G-ResNet18 &   84.71$\%$   \\ \hline                    
\end{tabular}
}
\end{minipage}
\end{table}

\section{Experiments}
\subsection{Datasets}
We validate the effectiveness of the proposed FSSD on five datasets. Firstly, we create  \emph{simple shape dataset} and create \emph{shape-AwA2 dataset} by transforming the popular animal dataset \emph{AwA2} \cite{banik2021novel}. Secondly, the public \emph{Caltech 101 shapes} dataset \cite{marlin2010inductive}, \emph{Workpiece} dataset \cite{lu2018viewpoint,gu2019cad,lu2020deep,wu2021domain} and \emph{Swedish leaf} dataset \cite{swedishleaf} are also used. Figure~\ref{fig:datdaset} shows some examples of the five datasets. Specifically, the simple shape dataset and shape-AwA2 are obtained as follows:
\begin{itemize}
    \item The simple shape dataset is developped using basic geometric elements such as lines, arcs, angles, and circles. It is created using PIL (Python Imaging Library) by randomly generating shapes with varying sizes, positions, orientations, and foreground/background colors. There are a total of 25 classes with 250,000 images in this dataset. 
    \item AwA2 dataset is a popular few-shot learning dataset. However, extracting shape features directly from color images 
    is challenging. Therefore, in this paper, a state-of-the-art salient object detector \cite{wu2019stacked} is employed to extract masks of animals from AwA2, creating the masks of AwA2 dataset specifically for shape classification research. This shape dataset consists of 50 classes with 37,322 images.
\end{itemize}
\subsection{Experiment Setup}
G-CNN with ResNet18 backbone is used for basic feature extraction. We use 5-way 1/5-shot setting with 15 query samples per class. We set the number of episodes to be 50,000 for training and 5,000 for testing. The Adam optimizer with learning rate 0.001 is used. The learning rate decays by 0.5 every 8,000 episodes. The classification accuracy is used to measure the performance of few-shot models, while PSNR (Peak Signal-to-Noise Ratio) and SSIM (Structural Similarity Index) are used to measure the performance of image reconstruction.

\subsection{Results of Few-shot Shape Recognition}
\begin{figure}[t]
\centering
\includegraphics[width=\linewidth]{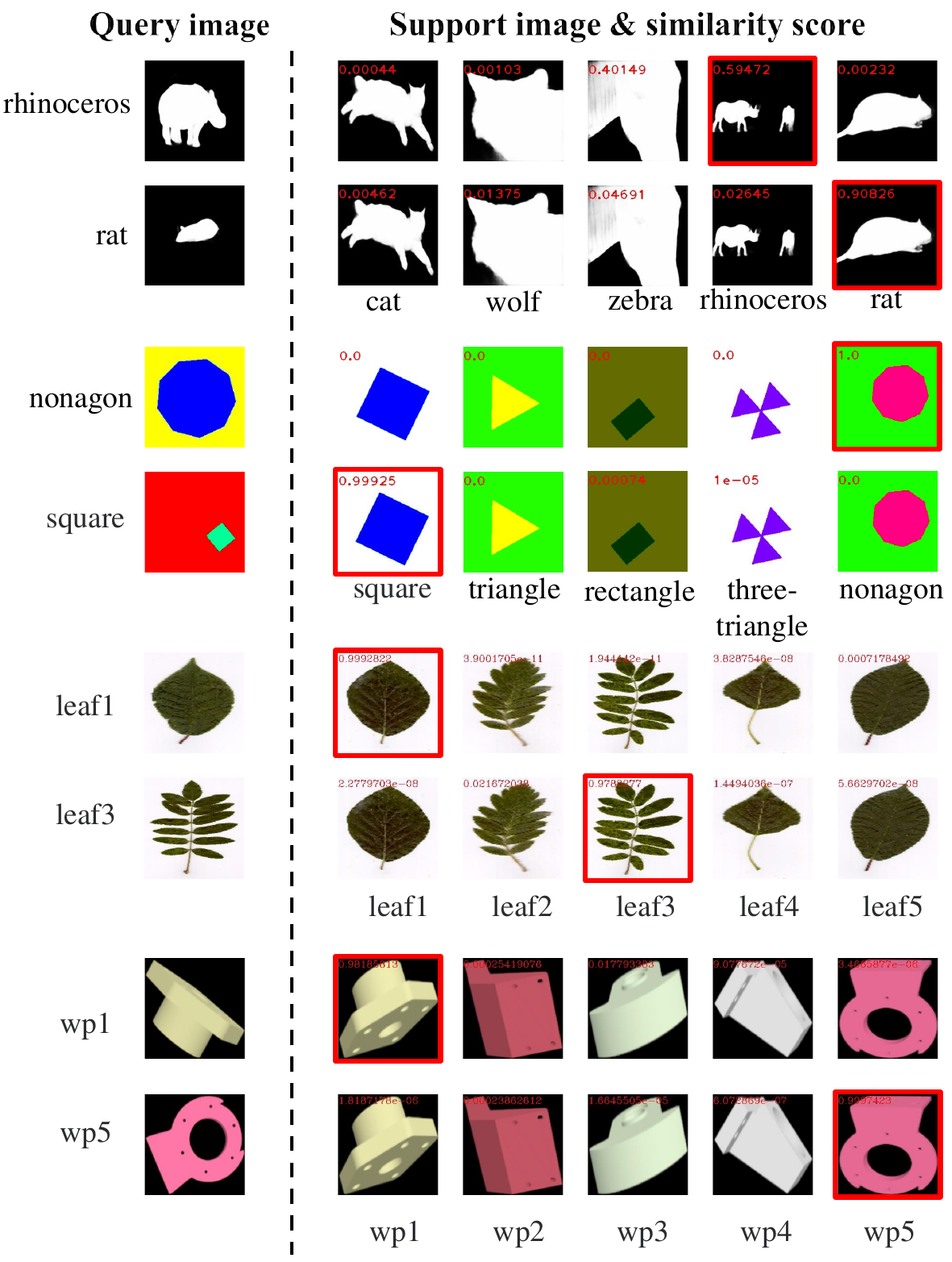} 
\caption{Visualizations of shape matching. Matched shapes are stressed by red frames. Similarity score is shown in red in the top-left corner. \emph{Best viewed in zoom.}}
\label{fig:matchPic}
\end{figure}

In this section, we compare the proposed FSSD with traditional shape descriptors such as Shape Context~\cite{belongie2002shape}, Fourier Descriptors~\cite{zahn1972fourier}, and Hu Moment~\cite{hu1962visual}, and classical few-shot learning methods such as ProtoNet~\cite{snell2017prototypical} and RelationNet~\cite{sung2018learning}. We compute 5-way 1/5-shot classification accuracy on five datasets by averaging 5,000 randomly generated episodes from the testing set. Table~\ref{table:Algorithm comparison} shows the superior performance of our method compared to other approaches on five datasets where the shape plays a prominent role. Our method demonstrates superior performance and efficacy in capturing shape information and outperforms the compared methods. Unlike us, methods such as CNNs struggle to capture shape information in 5-way 1-shot setting on the Workpiece and Swedish leaf datasets. While CNNs begin to improve performance with more templates being used, the achieved performance is still inferior to our method. 
Moreover, we visualize shape matching using our method on the shape-AwA2 dataset, the simple shape dataset, the Swedish leaf dataset, and the Workpiece dataset in Figure~\ref{fig:matchPic}, which shows that our model is able to successfully perform shape retrieval.

\subsection{Ablation Studies}
To our best knowledge, there is no existing research on shape-based neural networks for few-shot shape recognition. Therefore, we use state-of-the-art few-shot learning model \cite{sung2018learning} as the baseline and conduct ablation experiments on: (1) backbone types, (2) decoding methods, (3) similarity calculation methods, (4) attention types, and (5) model hyper-parameters.  



\vspace{0.1cm}
\noindent 
\textbf{Types of backbone. }
To examine the importance of G-CNN, the experiments are performed by only changing the backbone while keeping other modules the same. The impact of different backbones is shown in Table~\ref{table:table1}. On the simple shape dataset, we observe that the performance increases along with the model size. However, the margin becomes smaller between ResNet18 and ResNet50. If replacing ResNet18 by G-ResNet18 (G-CNN version), our model leads to significant improvements in performance (5.8\%). Moreover, using G-ResNet18 has very tiny overhead compared to the original CNN model (45.41M \vs 43M). Compared to ResNet50 (105.5M), G-ResNet18 (45.41M) enjoys much lower model capacity and higher performance, which shows the G-CNN extracts better shape representations.




\begin{table}[t]
\centering
\caption{
Results of different decoders using G-CNN.
}
\label{table:table3}
\newcommand{\tabincell}[2]{\begin{tabular}{@{}#1@{}}#2\end{tabular}}
\resizebox{\linewidth}{!}{
\begin{tabular}{llccccc}
\hline\noalign{\smallskip}
\multirow{2}*{Dataset} & \multirow{2}*{Decoder} & \multirow{2}*{Acc($\uparrow$)} & \multicolumn{2}{c}{Mask} & \multicolumn{2}{c}{Edge}\\
&&& SSIM($\uparrow$) & PSNR($\uparrow$) & SSIM($\uparrow$) & PSNR($\uparrow$)\\
\noalign{\smallskip}
\hline
\noalign{\smallskip}
\multirow{4}{*}{\tabincell{c}{Simple \\Shape}} & - & 89.01$\%$  & - & - & - & -\\
& Mask & 89.09$\%$ & 0.8636 & 40.01 & - & -\\
& Edge & 88.90$\%$ & - & - & 0.8943 & 38.41\\
&\cellcolor{green!20}Mask+Edge & \cellcolor{green!20}\textbf{91.02$\%$} & \cellcolor{green!20}0.8926 & \cellcolor{green!20}40.01 & \cellcolor{green!20}0.9131 & \cellcolor{green!20}38.81\\
\hline
\end{tabular}
}
\end{table}

\begin{table*}[t]
\centering
\caption{
Comparison of different distances and similarity metrics.
}
\label{table:table4}
\resizebox{0.7\linewidth}{!}{
\begin{tabular}{llccccccc}
\hline\noalign{\smallskip}
\multirow{2}*{Backbone} & \multirow{2}*{Decoder} & \multirow{2}*{Method} & \multirow{2}*{Acc($\uparrow$)} & \multicolumn{2}{c}{Mask} & \multicolumn{2}{c}{Edge}\\
&&&& SSIM($\uparrow$) & PSNR($\uparrow$) & SSIM($\uparrow$) & PSNR($\uparrow$)\\
\noalign{\smallskip}
\hline
\noalign{\smallskip}
\multirow{2}{*}{G-CNN} & \multirow{2}{*}{Mask+Edge} & Relation module & 66.23$\%$ & 0.8815 & 42.77 & 0.9455 & 39.67\\
& & \cellcolor{green!20} Cosine similarity & \cellcolor{green!20}\textbf{91.02}$\%$ & \cellcolor{green!20}0.8926 & \cellcolor{green!20}40.01 & \cellcolor{green!20}0.9131 & \cellcolor{green!20}38.81\\
\hline
\end{tabular}
}
\end{table*}

\vspace{0.1cm}
\noindent 
\textbf{Different decoder methods. }
\label{sec:Different decoder methods}
To investigate the influence of shape mask/edge decoders, we only vary the decoders while the other modules are same. Table \ref{table:table3} shows the results of different decoding methods. One can see how both decoders contribute to the classification and image reconstruction measured by SSIM and PSNR. Using both decoders simultaneously provides superior performance compared to each single one. This also implies that the mask and edge groundtruth are complementary; the former provides the global information while the latter captures the edge information. 




\vspace{0.1cm}
\noindent 
\textbf{Calculation of similarity. }
This experiment compares cosine similarity \vs the relation module popular in few-shot learning \cite{sung2018learning}. The relation module learns a good ``metric'' in a data-driven way and enjoys better performance than conventional distances such as the Euclidean or cosine distance. Table \ref{table:table4} shows that using the cosine similarity in few-shot shape recognition is  reasonable. We believe this is mainly due to the superior shape features learned through our model, thus, the performance depends less on the distance metric. 


\begin{table*}[t]
\centering
\caption{
Comparison of H-MCA and S-MCA.
}
\label{table:table6}
\resizebox{0.7\linewidth}{!}{
\begin{tabular}{llcccccc}
\hline\noalign{\smallskip}
\multirow{2}*{Backbone} & \multirow{2}*{Attention} & \multirow{2}*{Acc($\uparrow$)} & \multicolumn{2}{c}{Mask} & \multicolumn{2}{c}{Edge}\\
&&& SSIM($\uparrow$) & PSNR($\uparrow$) & SSIM($\uparrow$) & PSNR($\uparrow$)\\
\noalign{\smallskip}
\hline
\noalign{\smallskip}
\multirow{3}{*}{G-CNN} & S-MCA & 90.13$\%$  & 0.9180 & 42.79 & 0.9395 & 39.39\\
& H-MCA & 89.21$\%$ & 0.8286 & 37.32 & 0.8854 & 37.12\\
& \cellcolor{green!20} Dual attention & \cellcolor{green!20}\textbf{91.02}$\%$ & \cellcolor{green!20}0.8926 & \cellcolor{green!20}40.01 & \cellcolor{green!20}0.9131 & \cellcolor{green!20}38.81\\
\hline
\end{tabular}
}
\end{table*}

\vspace{0.1cm}
\noindent 
\textbf{Attention architecture.}
This experiment uses G-ResNet18 as the backbone, a pair-wise decoder, and the cosine similarity while exploring the efficacy of different attention architectures. As shown in Table~\ref{table:table6}, a standalone S-MCA achieves higher classification accuracy and better image reconstruction quality than H-MCA. Nonetheless, S-MCA is not meant to discover the common primitives across different heads and leads to weak interpretations in visualization. On the other hand, H-MCA provides comparable performance in accuracy but shows relatively weak performance in reconstruction, as indicated in the methodology section. The dual attention architecture enables the integration of both modules to complement each other. As a result, the combined model obtains the best classification results and better reconstruction than the single H-MCA.


\begin{figure}
\centering
\includegraphics[width=\linewidth]{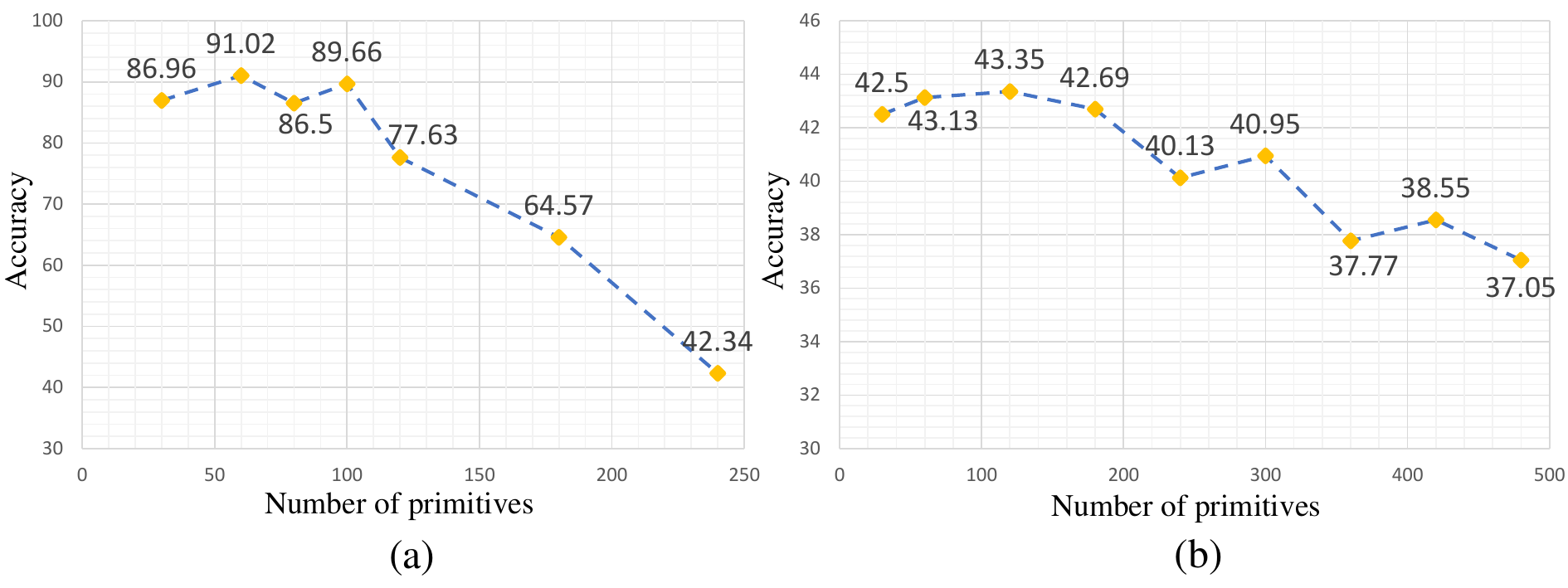} 
\caption{
(a) Shape recognition performance using 30, 60, 80, 100, 120, 180, and 240 shape primitives on the simple shape dataset. (b) Same experiments tested on the shape-AwA2 dataset.}
\label{fig:primitiveNum}
\end{figure}

\vspace{0.1cm}
\noindent 
\textbf{Numbers of shape primitives. }
Figures~\ref{fig:primitiveNum}(a) and (b) show the results with different numbers of shape primitives on the simple shape dataset and shape-AwA2 dataset, respectively. While more shape primitives seem to provide the enriched basis for reconstruction, they may also contain redundant information, leading to the overfitting issue, which can be observed in reconstruction-based feature learning. One may observe in both figures that the performance of the network did not improve with increasing numbers of primitives. When the number of primitives exceeds a specific point, the classification performance drops quickly. Therefore, maintaining a moderate number of shape primitives is critical. According to  Figures~\ref{fig:primitiveNum}(a) and (b), for the optimal performance, we set the number of primitives to be 60 for the simple shape dataset and 120 for the shape-AwA2 dataset \textit{by default} throughout all  experiments. 

\begin{figure}[t]
\centering
\includegraphics[width=.8\linewidth]{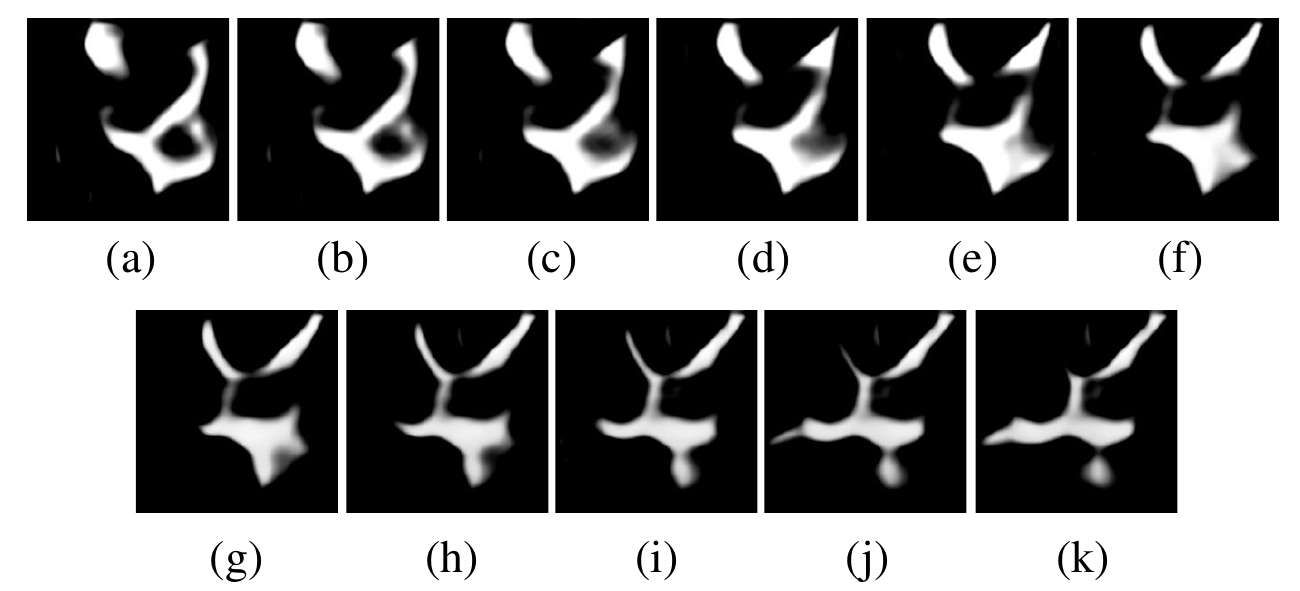}
\caption{
Images (a) and (k)  visualize primitives $\boldsymbol{\phi}_1$ and $\boldsymbol{\phi}_2$, and (b)-(j) visualize the feature vectors interpolated by $\boldsymbol{\phi}_1$ and $\boldsymbol{\phi}_2$.}
\label{fig:interpolation}
\end{figure}
\begin{figure}[t]
\centering
\includegraphics[width=.8\linewidth]{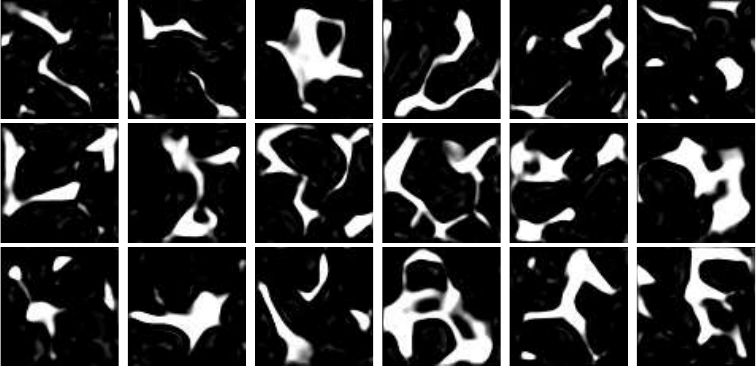}
\caption{
Partial visualization of primitives on the simple shape dataset. The number of shape primitives is set to 60.}
\label{fig:primitiveVisual}
\end{figure}

\vspace{0.1cm}
\noindent 
\textbf{Visualization of shape primitives. }
\label{sec:Visualization of shape primitives}
To investigate the visualization of shape primitives, we selected two primitives $\boldsymbol{\phi}_1, \boldsymbol{\phi}_2$, and performed interpolation on them as follows:
\begin{equation}
  \begin{aligned}
    \mathbf{\phi}_{i}^{inter} &= \alpha \times ~\boldsymbol{\phi}_{1} + \left( {1 - ~\alpha} \right) \times ~\boldsymbol{\phi}_{2}~ \\
    \alpha &= i \times 0.1,~~i = 0,\ldots,~10. \\
\end{aligned}
\label{equ:14}
\end{equation}
Through the interpolation of $\boldsymbol{\phi}_1$ and $\boldsymbol{\phi}_2$, 11 interpolated feature vectors were obtained. These feature vectors were then fed into the decoder to decode them into corresponding masks, as shown in Figure~\ref{fig:interpolation}. The figure shows that the reconstructed images corresponding to the interpolated features are continuously changing. This demonstrates that the decoded images corresponding to the linearly combined primitives are obtained by continuous variation in the reconstruction space. Figure~\ref{fig:primitiveVisual} shows the visualization results of some primitives of the simple shape dataset when the number of primitives is set to 60. From the results, it is not straightforward to link the visualization of primitives to a real yet complete shape. However, both the theoretical outcomes in Eq.~\ref{equ:9} and the decoding of the reconstructed interpolated features in Figure~\ref{fig:interpolation} demonstrate that these primitives can approximate the original shape well.

\renewcommand{\dblfloatpagefraction}{.5}

\begin{figure}[t]
\centering
\includegraphics[width=.786\linewidth]{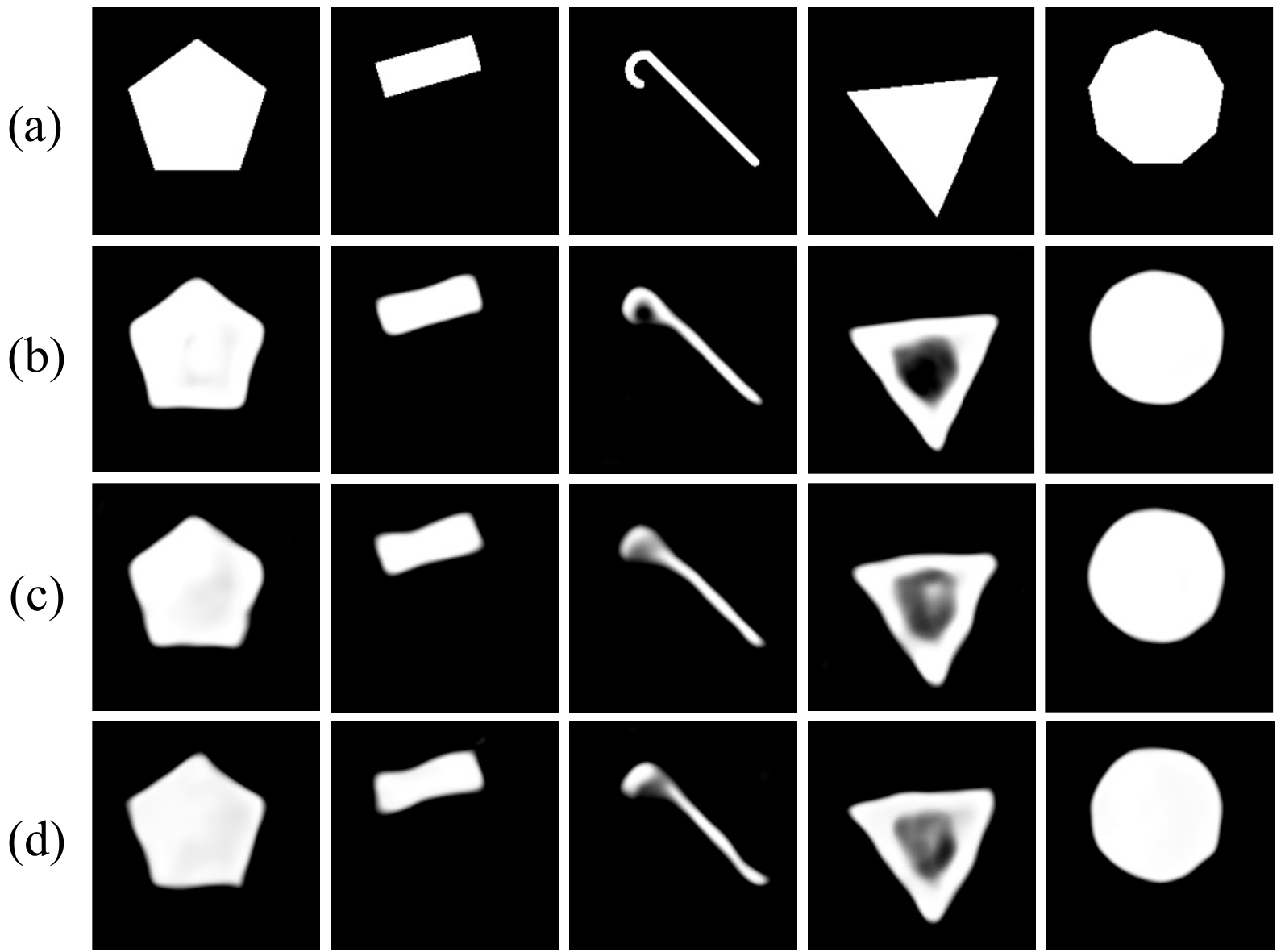}
\caption{Visualization of reconstructed images: (a) ground truth; (b), (c), (d) are the decoding results of $\mathbf{Q}, \mathbf{Q}'_H$ and $\mathbf{Q}'$.}
\label{fig:decoderImage}
\vspace{-10pt}
\end{figure}

\noindent 
\textbf{Visualization of reconstructions.} 
Figure~\ref{fig:decoderImage} shows the visualization results of $\mathbf{Q}, \mathbf{Q}'_H$ and $\mathbf{Q}'$ on the test dataset through the decoder using the mask branch. One can see that for shapes not included in the training dataset, our model can still leverage primitives to reconstruct them in a precise manner. These results demonstrate that dual attention is needed for faithful and interpretable reconstruction. For instance, the pentagon in Figures~\ref{fig:decoderImage}(a), (c) and (d) represent  ground truth, results by H-MCA and H-MCA+S-MCA, respectively. One may see that H-MCA is not sensitive to obtuse angles, and the decoder tends to reconstruct obtuse angles as arcs. However, the integration with S-MCA can do a better job of reconstructing the obtuse angles.

\vspace{-0.2cm}

\section{Conclusions}
We have proposed  to incorporate shape properties in the network architecture so that the network can perform efficient few-shot shape learning. Our proposed FSSD  replaces the CNN with a G-CNN to enable the network obtain robustness to geometric variations of object poses. An image reconstruction module is used and a pair-wise decoder architecture to ensure the output image focuses on the edge information of the target. A dual attention architecture is used to implement shape primitives learning, and the shape primitives are used to reconstruct new features to approximate the original features. Our FSSD is robust to shape transformations and has the ability to reconstruct shapes due to improved shape modeling. The extensive experiments on few-shot shape recognition show the effectiveness of the proposed approach. Despite the learned geometric primitives differ from human-made geometric primitives (\eg, lines, small arcs), we hope our work may inspire the future research in this field.
 

\balance
{\small
\bibliographystyle{ieee_fullname}
\bibliography{refs}
}

\end{document}